\documentclass[11pt]{article}

\usepackage[margin=0.9in]{geometry}
\usepackage{amsmath,amssymb}
\usepackage{booktabs}
\usepackage{array}
\usepackage{graphicx}
\usepackage{flafter}
\usepackage{placeins}
\usepackage{needspace}
\usepackage{microtype}
\usepackage[authoryear,round]{natbib}
\usepackage{xcolor}
\usepackage{hyperref}
\usepackage{url}
\hypersetup{
  colorlinks=true,
  linkcolor=blue,
  citecolor=blue,
  urlcolor=blue,
  pdftitle={SteerCheck: Attribution Specificity and Alignment Leakage in Activation-Steering Audits},
  pdfauthor={Daming Luo, Christy Liang, Junyu Xuan}
}

\newcommand{\kl}{\mathrm{KL}}
\newcommand{\method}{\textsc{SteerCheck}}
\newcommand{\vcaa}{v_{\mathrm{CAA}}}
\newcommand{\vshuf}{v_{\mathrm{shuf}}}
\newcommand{\sd}{\mathrm{sd}}
\newcommand{\rms}{\mathrm{rms}}

\title{\method: Attribution Specificity and Alignment Leakage\\ in Activation-Steering Audits}

\author{
  Daming Luo \quad Christy Liang \quad Junyu Xuan\\[3pt]
  University of Technology Sydney\\[3pt]
  \texttt{daming.luo@student.uts.edu.au}\\
  \texttt{Jie.Liang@uts.edu.au} \quad \texttt{Junyu.Xuan@uts.edu.au}
}

\date{}

\begin{document}
\maketitle

\begin{abstract}
Activation steering can change behaviour without establishing that the effect is
specific to the intended concept.  We introduce \method{}, a preregistered
attribution audit that matches off-target KL and separates mean, protected-tail,
polarity, transfer, and semantic claims.  Exact replay of 960 Qwen3-14B
interventions reveals complementary limits of common controls: isotropic
directions occupy a narrow near-orthogonal region, whereas sign-randomized
same-construction directions often retain substantial target alignment.  Effect
is strongly associated with signed cosine within the sign-randomized family
($\rho=.94$); $25.3\%$ of its draws exceed cosine $.5$, and every draw exceeding
the observed mean effect has cosine above $.80$.  This alignment leakage does
not by itself invalidate a conditional randomization test; it limits what the
comparator can distinguish and motivates reporting exchangeability assumptions,
a construction diagnostic $A$, and the empirical cosine distribution.  The
primary Qwen complete gate remains negative because the protected tail fails all
families.  On independent data, continuous margin transfers only in Qwen and
accuracy transfers in no selected cell.  Prospectively registered language
controls pass the complete gate in Qwen and DeepSeek, while a passing DeepSeek
detox comparator rules out categorical separation; all nominal passes are
sensitive to $\Gamma=1.10$.  Frozen three-rater open-generation evaluation
supports factual correction in DeepSeek but not Qwen; the automatic judge fails
calibration (macro-F1 $.562$), so null-wide semantic results remain descriptive.
\method{} makes these conditional and mixed conclusions auditable.
\end{abstract}

\section{Introduction}

Activation steering modifies a language model at inference time by adding a
vector to an intermediate representation.  Contrastive mean differences, learned
head directions, and latent control vectors can alter truthfulness, style,
refusal, and other behaviours without updating model weights
\citep{subramani2022latent,li2023iti,turner2023activation,rimsky2024caa,
zou2023representation}.  A behavioural change alone, however, does not identify
its cause: the same result may arise from the intended concept, a broadly
disruptive direction, a construction artifact, or a tendency to oppose the user.

The standard evidence for attribution is a random-direction control.  A steering
vector is compared with directions of matched Euclidean norm; if it moves the
behaviour score further, the direction is often described as concept-specific.
That comparison has a narrow estimand: in a high-dimensional space, isotropic
draws probe the near-orthogonal region.  A stronger same-construction comparator
rebuilds the direction after independently reversing paired contrasts, but its
randomization interpretation is conditional on within-pair orientation
exchangeability and it need not be alignment-free.

We measure both limitations at a matched functional budget.  Exact reconstruction
of 320 isotropic, 320 PCA-subspace, and 320 sign-randomized directions recovers
their signed cosines with CAA.  Isotropic cosines have standard deviation $.014$;
the sign-randomized family has standard deviation $.577$, with $25.3\%$ of draws
above cosine $.5$.  Within that family, signed cosine and measured effect have
Spearman $\rho=.94$.  The result is not a new validity test: exchangeability
determines randomization validity.  It is an alignment-leakage and
discriminability diagnostic showing that label randomization can retain much of
the observed direction after normalization and KL matching.

Practically, an audit should match a functional budget, show the effect--alignment
profile rather than only a threshold crossing, report the assumptions and target
alignment of any same-construction comparator, and decompose ``steering worked''
into separately testable claims.  Appendix~\ref{app:protocol} gives the six-step
protocol and the failure each step catches.

\method{} implements these as a preregistered protocol with matched off-target
KL, three null families, an intersection--union specificity gate, polarity
mirroring, and prospective positive controls.  Statistical gates, Monte Carlo
budgets, split access, and stop rules were frozen before their corresponding
outputs were observed, and every failed, aborted, or superseded run is retained
rather than replaced.  Section~\ref{sec:positive} shows the gate is passable;
Sections~\ref{sec:null} and \ref{sec:dose} report the alignment diagnostics;
Section~\ref{sec:other} reports the transfer and polarity results.

The preregistered Qwen question was whether CAA exceeds all three families at
Holm $\alpha=.01$.  The complete answer remains no: the strict tail fails all
three families, and the mean also fails against sign randomization.  The cosine
analysis is post-hoc, exactly reproducible from the frozen seed, and changes the
scientific interpretation of the comparator without changing the frozen verdict.

Our contributions are a measured effect--alignment profile over 960 matched-KL
interventions; a construction diagnostic $A$ and a 21-cell measurement of
alignment retained by sign randomization; prospective evidence that the complete
gate is passable in two model families; and a frozen human-calibrated
open-generation test that separates a DeepSeek pass, a Qwen non-pass, and an
automatic-judge calibration failure.  The primary
evidence concerns one anti-sycophancy/truth-consistency behaviour, CAA as the
real-direction construction, and one selected layer per model;
Section~\ref{sec:limits} states the boundary.

\begin{figure}[t]
\centering
\includegraphics[width=\linewidth,height=2.75in,keepaspectratio]{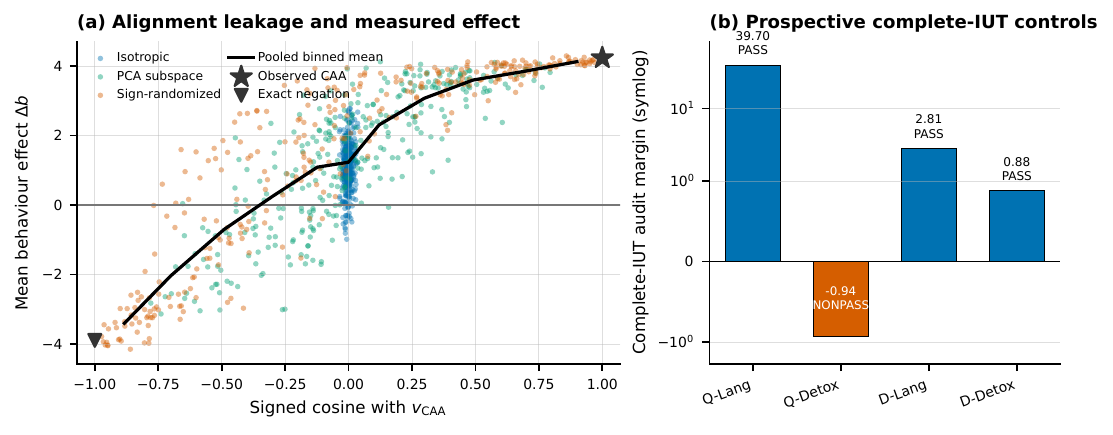}
\caption{Two complementary diagnostics.  (a) Post-hoc exact replay of 960
matched-KL Qwen3-14B layer-28 interventions.  Points are measured forward
passes; the black curve is a descriptive pooled binned mean, while the
family-stratified associations are $\rho=.938$ for sign randomization and $.843$
for PCA).  Broad sign-randomized cosine support is target-alignment leakage, not
by itself evidence that randomization is invalid.  (b) Prospective P2G
complete-IUT margins: both language controls and DeepSeek Detox pass; Qwen Detox
does not.  Q-Lang/Q-Detox and D-Lang/D-Detox denote Qwen and DeepSeek language
controls/comparators.  P2G uses a separately frozen balanced block-Rademacher
design, not the iid S3 design in panel (a).}
\label{fig:dose}
\end{figure}

\section{Related work}

\paragraph{Activation steering.}
Latent steering vectors have been extracted from pretrained decoders for
controlled generation \citep{subramani2022latent}.  Activation Addition and CAA
form directions from contrastive activations and add them during inference
\citep{turner2023activation,rimsky2024caa}.  ITI learns and intervenes on
truth-related attention-head directions \citep{li2023iti}, while representation
engineering treats population-level representations as a general interface for
monitoring and control \citep{zou2023representation}.  These works establish
that activation interventions change outputs.  We ask what control and
calibration are required to attribute that change to a labelled direction.

\paragraph{Evaluation reliability.}
Prior frameworks call for likelihood-aware metrics, downstream-like contexts,
standardized comparisons, and explicit baselines \citep{pres2024reliable}.
Steering vectors are brittle under prompt shifts and vary across models and
methods \citep{tan2024reliability,dasilva2025steering}; geometry and coherence
predict further reliability failures \citep{braun2025unreliability}; changing
only the activation source can change steering success \citep{ye2026sources};
open-ended evaluation introduces coherence and threat-model requirements absent
from forced choice \citep{herbster2026openended}; and
\citet{goyal2026specificity} study behavioural specificity across general,
control, and shifted conditions.

\paragraph{What is new here.}
This literature argues that steering evaluations are unreliable; it does not
measure the association between effect and direction alignment at a fixed
functional budget, nor quantify how much target alignment a same-construction
sign-randomization comparator retains.  Where
\citet{ye2026sources} change the activation source, we hold it fixed; where
\citet{goyal2026specificity} target \emph{behavioural} specificity across
conditions, we target \emph{attribution} specificity at a fixed source and
operating point.  Appendix~\ref{app:diag}, Table~\ref{tab:related} locates our
contribution as a combination of controls rather than claiming any one component
is individually new.

\section{Matched-budget steering audits}
\label{sec:method}

For paired construction examples, let $h_\ell(x_i^+)$ and $h_\ell(x_i^-)$ be the
last-token residual activations for the positive and negative members of pair
$i$.  CAA forms the normalized mean contrast and adds it after the prompt:
\begin{equation}
d_i=h_\ell(x_i^+)-h_\ell(x_i^-),\qquad
\vcaa=\frac{\bar d}{\|\bar d\|_2},\quad
\bar d=\frac1n\sum_{i=1}^{n}d_i,\qquad
h'_{\ell,t}=h_{\ell,t}+cv\ \ (t>t_{\mathrm{prompt}}).
\end{equation}
On a neutral prompt bank $\mathcal{N}$ disjoint from the behavioural data, the
off-target budget is
\begin{equation}
\kappa=\mathbb{E}_{x\sim\mathcal{N}}
\left[\kl\left(p_{\theta}(\cdot\mid x)\,\|\,p_{\theta,cv}(\cdot\mid x)\right)\right].
\end{equation}
The expectation uses a frozen 16-token horizon; all confirmatory directions are
matched to $\kappa\in[0.09,0.11]$.

\paragraph{Why the budget is KL rather than norm.}
For small interventions,
\begin{equation}
\kappa\approx\tfrac12 c^2v^\top Fv,
\qquad
\Delta s\approx\sqrt{2\kappa}\,\frac{v^\top\nabla s}{\sqrt{v^\top Fv}},
\label{eq:quad}
\end{equation}
where $F$ is the local Fisher information and $s$ is a behaviour score.
Matching $\kappa$ fixes the perturbation's Fisher norm and leaves alignment with
the behaviour gradient free---the quantity a specificity audit must isolate.
Matching $\|v\|_2$ instead leaves $v^\top Fv$ free, so norm-matched directions
need not be comparably dosed.  The matched budget is conditional on the
reference-prompt distribution, token horizon, and model setting.

For each direction, a bracketed search selects $c$ using only the neutral-KL
bank; behaviour scores, answer accuracy, and confirmation outputs never tune it.
Direction hashes are written before any confirmation output is accessed.
Steering is applied at the frozen layer to post-prompt token positions.  The
negative-direction gate uses the exact negation of the observed vector.

\subsection{Null families}

We compare the observed direction with three families: \textbf{isotropic}
random directions; directions inside the top-eight activation \textbf{PCA
subspace}; and \textbf{sign-randomized} directions.  Writing $U_k$ for the
top-$k$ PCA basis and $\epsilon_i^{(b)}\overset{\mathrm{iid}}{\sim}
\operatorname{Rad}(1/2)$,
\begin{align}
v_{\mathrm{iso}}^{(b)}&=\frac{z_b}{\|z_b\|_2},\quad
z_b\sim\mathcal N(0,I_d),
&v_{\mathrm{PCA}}^{(b)}&=\frac{U_ka_b}{\|U_ka_b\|_2},\quad
a_b\sim\mathcal N(0,I_k),\\
\vshuf^{(b)}&=\frac1n\sum_{i=1}^{n}\epsilon_i^{(b)}d_i.&&
\end{align}
The isotropic family is a geometric null: it preserves nothing but scale.  The
PCA family preserves the dominant activation subspace.  The sign-randomized
family preserves paired rows, activation geometry, estimator, and scale, and
attempts to remove the coherent label association while retaining the
construction---which is why it was preregistered as the strongest of the three.
Its randomization inference is conditional on within-pair sign exchangeability.
It need not, however, be geometrically unrelated to the observed direction.
Section~\ref{sec:null} measures that retained alignment and clarifies what the
comparator can distinguish; alignment overlap alone is not a proof that the
randomization model is invalid.

\subsection{Endpoints, gates, and what a non-pass means}
\label{sec:gates}

The discovery audit uses a continuous forced-choice logit contrast $b$.  The
independent-bank study reports both the mean margin change and the change in
forced-choice accuracy.  Within each family, the specificity gate is
intersection--union: both a mean statistic and a strict lower-tail statistic
must pass Holm-corrected $\alpha=0.01$ \citep{holm1979}.  Null ordering and
signed direction transport are separate preregistered gates, and models are
never pooled.

For 128 item-level effects the tail statistic $z_7$ is the seventh-smallest
effect, approximating the empirical fifth percentile; both statistics are
compared to each family's Monte Carlo distribution and Holm-adjusted within
statistic.  Appendix~\ref{app:protocol} gives the estimator definitions, the
finite-sample $p$-value, and the null-ordering statistic $D_{95}$.

\paragraph{How to read a non-pass.}
An intersection--union test passes only if every component passes, so a non-pass
is only as interpretable as the weakest component.  Two limits apply throughout.
First, the protected-tail component's power against any alternative is
uncalibrated: our implant ladder calibrates the mean component only.  Second,
a family can be exchangeable yet retain substantial alignment with the target,
which limits the alternatives that its comparison can discriminate.  We
therefore report component-wise results, name the responsible family, and keep
randomization validity (exchangeability) separate from geometric
discriminability (alignment leakage).

We did not redefine any gate after seeing results.  The intersection--union
definition, the tail order statistic, and $\alpha$ were frozen before
confirmatory access, and the preregistration forbids re-tuning to flip a
verdict.  Section~\ref{sec:null} refines what one comparator can establish,
without changing a threshold or the original verdict retained in
Appendix~\ref{app:ledger}.

\subsection{Prospective positive-control design}

The language positive control and detox copy-controlled comparator were
prespecified before confirmation access; geometry does not choose behaviours.
Construction-only banks select one layer per fixed model--behaviour cell under
the lower-layer tie rule $\ell^*=\min\arg\max_\ell C_\ell$, and the reserve bank
remains sealed.  Synthetic-only power and throughput analyses selected the
512-item configuration with $m=512$, $B=999$, and 256 unrelated neutral-KL
prompts; the protected statistic is $\delta_{(26)}$.  Each of the four cells
independently applies the strict Holm $\alpha<.01$ complete IUT against
isotropic, PCA-subspace, and separately frozen balanced block-Rademacher nulls,
which assign exactly eight positive and eight negative signs within each block
of~16.  Models and behaviours are not pooled.

\section{The gate is passable: prospective positive controls}
\label{sec:positive}

\emph{Chronology.}  This study was prespecified and executed \emph{after} the
audit in Section~\ref{sec:null}.  We present it first because a negative audit
is uninterpretable until the gate is shown to be passable.
Appendix~\ref{app:ledger} preserves the actual execution order, and no result
here was used to modify any gate applied elsewhere.

\begin{table}[t]
\caption{Prospective positive-control cells.  Language is the FLORES
English-to-French continuation control; Detox is the copy-controlled semantic
comparator.  Decisions are cell-specific complete IUTs at strict Holm
$\alpha=.01$; non-pass is not equivalence to the null.}
\label{tab:p2g}
\centering
\small
\begin{tabular}{llrrrrrc}
\toprule
Model & Behaviour & Layer & LOO & Mean & Tail & Margin & Complete IUT \\
\midrule
Qwen3-14B        & Language & 20 & 0.708 & 12.5 & 7.17   & 39.70 & \textbf{PASS} \\
Qwen3-14B        & Detox    & 28 & 0.403 & 1.12 & $-0.285$ & $-0.94$ & non-pass \\
DeepSeek-V2-Lite & Language & 24 & 0.717 & 1.23 & 0.534  & 2.81  & \textbf{PASS} \\
DeepSeek-V2-Lite & Detox    & 18 & 0.402 & 2.65 & 0.611  & 0.88  & PASS \\
\bottomrule
\end{tabular}
\end{table}

Three facts follow.  First, the complete gate---including the protected-tail
component---is passable, in two model families, under the same matched-KL
protocol and the same three comparator roles used elsewhere.  Its structured
family is a separately frozen balanced block-Rademacher design; the iid
alignment diagnostic in Section~\ref{sec:null} does not transfer to it without a
separate analysis.  Second, the predicted
within-model ordering holds: in each model the language-control margin exceeds
the detox comparator's.  Third, ordering is not categorical separation: Qwen
Detox does not pass while DeepSeek Detox does, so construction-only coherence
ranks cells without partitioning them.  All four ledgers contain 2{,}999 unique
successful keys with no reserve-bank access, every direction independently
matched to the same off-target KL interval; a disclosed fixed-horizon
implementation correction occurred after authorized confirmation access but
before any formal result event (Appendix~\ref{app:integrity}).

\paragraph{Boundary.}
The three nominally passing cells pass under the frozen randomization model, but
none remains robust under a conservative $\Gamma=1.10$ sensitivity envelope.
Exact negation and sensitivity are secondary diagnostics and cannot rescue a
primary non-pass.  This study calibrates the frozen CAA constructions and does
not validate all steering methods or direction constructions.

\section{Target-direction leakage limits what a matched comparator can distinguish}
\label{sec:null}

We audit CAA on Qwen3-14B layer 28 with 128 anti-sycophancy rows
\citep{rimsky2024caa,yang2025qwen3}.  An initial $B=100$ screen could not
resolve a three-family Holm test at $\alpha=.01$ (minimum adjusted $p$ is
$3/101\approx.0297$); a synthetic power gate required $B\ge299$, and the
confirmatory audit uses $B=320$.

\begin{table}[t]
\caption{Confirmatory audit at $B=320$, with each family's measured alignment to
the direction it is a null for.  Observed mean $\Delta b=4.2321$; a family passes
the mean component when Holm $p<.01$.  The left block is what the audit sees; the
right block is why.}
\label{tab:sc1}
\centering
\scriptsize
\setlength{\tabcolsep}{3pt}
\begin{tabular}{lrrrl@{\hspace{1.2em}}rrrr}
\toprule
& \multicolumn{4}{c}{effect at matched KL} & \multicolumn{4}{c}{alignment with $\vcaa$} \\
\cmidrule(lr){2-5}\cmidrule(lr){6-9}
Null family & $q_{95}$ & Exceed. & Holm $p$ & Mean comp. &
$\sd\cos$ & $q_{05}$ & $q_{95}$ & $P(\cos\!\ge\!.5)$ \\
\midrule
Isotropic       & 2.349 & 0/320 & .00935 & pass & .0141 & $-.022$ & $+.024$ & .000 \\
PCA subspace    & 3.841 & 0/320 & .00935 & pass & .3376 & $-.554$ & $+.574$ & .078 \\
Sign-randomized & 4.115 & 4/320 & .01558 & fail & \textbf{.5769} & $\mathbf{-.873}$ & $\mathbf{+.875}$ & \textbf{.253} \\
\bottomrule
\end{tabular}
\end{table}

Table~\ref{tab:sc1} gives two different facts that must not be conflated.  The
frozen conditional randomization comparison fails its mean gate, while the
same draws retain much more target-direction alignment than the isotropic
family.  The former remains the preregistered decision under within-pair sign
exchangeability; the latter measures the alternatives that this comparator is
able to separate from the observed construction.

\paragraph{The randomized directions retain target alignment.}
The sign-randomized directions come from a single frozen random stream, so they
reconstruct exactly; validating against each draw's recorded perturbation norm
matches to $1.7\times10^{-16}$ relative error for all 320 draws.  Their cosines
with $\vcaa$ are then immediate, and they appear in the right-hand block of
Table~\ref{tab:sc1}.

A quarter of the sign-randomized draws are more than half aligned with $\vcaa$;
one in twenty is more than $90\%$ aligned.  All four draws that exceed the
observed effect have cosines of $.9286$, $.8664$, $.8584$, and $.8044$---they
are close in direction to the observed vector.  This geometry explains why the
structured reference distribution approaches the observed effect.  It does not
by itself invalidate the Holm $p=.01558$: that $p$-value is conditional on the
frozen exchangeability model.  It does show that the comparison is not a test
against alignment-free alternatives and therefore supports a narrower
specificity claim.

\paragraph{Why, and when it happens.}
The randomized vector is mean-zero---$\mathbb E[\vshuf]=0$---but its normalized
\emph{alignment} is governed by a different quantity.  Write
$u=\bar d/\|\bar d\|_2$ and $p_i=d_i^\top u$.  The component of $\vshuf$ along
$u$ is $\frac1n\sum_i\epsilon_ip_i$, with standard deviation $\rms(p)/\sqrt n$,
while $\|\vshuf\|$ concentrates at $\rms(\|d_i\|)/\sqrt n$.  The $\sqrt n$
cancels, leaving
\begin{equation}
\sd\big(\cos(\vshuf,v_{\mathrm{obs}})\big)\approx A,
\qquad
A=\frac{\rms_i\big(d_i^\top u\big)}{\rms_i\big(\|d_i\|_2\big)} .
\label{eq:A}
\end{equation}
$A$ is the root-mean-square fraction of an \emph{individual} pair contrast's
energy that lies along the mean direction---not a property of the mean's
magnitude.  In this first-order approximation the $\sqrt n$ factors cancel, so
increasing the number of pairs does not automatically shrink the normalized
alignment spread.  $A$ is therefore an alignment-concentration diagnostic, not
a validity threshold: randomization validity still depends on exchangeability.
For the Qwen3-14B layer-28 cell, $A=.606$ against the measured
$\sd(\cos)=.577$.

\paragraph{The pattern recurs across the available banks.}
Because $\cos(\vshuf,v_{\mathrm{obs}})$ depends only on the stored paired
activation bank, it is computable without any forward pass.  We evaluated every
eligible construction bank we hold---three construction banks over seven layers
each: Qwen3-14B anti-sycophancy, Qwen3-14B benign-compliance, and Qwen2.5-7B
anti-sycophancy \citep{qwen2024qwen25}---by simulating 4{,}000 sign-randomized
draws per cell.  These are two related Qwen model series, not independent
model-family replication.

\begin{table}[ht]
\caption{Alignment leakage across 21 constructions.  $A$ is the predicted, and
$\sd(\cos)$ the simulated, standard deviation of the randomized direction's
alignment with the observed direction.  No pass/fail cutoff was preregistered.}
\label{tab:transfer}
\centering
\small
\begin{tabular}{llcccc}
\toprule
Behaviour & Model & Layers & $A$ range & $\sd(\cos)$ range &
$P(\cos\ge.5)$ range \\
\midrule
anti-sycophancy   & Qwen3-14B  & 8--32 & .424--.977 & .440--.921 & .154--.480 \\
benign-compliance & Qwen3-14B  & 8--32 & .777--.972 & .640--.870 & .321--.456 \\
anti-sycophancy   & Qwen2.5-7B & 6--22 & .436--.983 & .454--.917 & .154--.471 \\
\bottomrule
\end{tabular}
\end{table}

$A$ predicts the simulated $\sd(\cos)$ with Pearson $r=.979$ and mean absolute
error $.075$ across the 21 cells; every measured $\sd(\cos)$ is at least $.440$.
This is a continuous diagnostic with no confirmatory threshold.  The result
shows substantial retained alignment across all three available banks and
layers; it does not establish a universal property of contrastive
mean-difference constructions or an exchangeability failure.

\paragraph{The reporting requirement.}
For a construction-matched comparator, we recommend publishing $A$ and the
empirical distribution of $\cos(v_{\text{null}},v_{\text{obs}})$.  Both are one
line of linear algebra on the activation bank and need no forward pass.  The
randomization $p$-value can then be read together with what the comparator
actually removes and retains, instead of being mistaken for an alignment-free
test.

\paragraph{What survives.}
The isotropic family provides a near-orthogonal geometric baseline
($\sd(\cos)=.014$), and CAA clears it at Holm $p=.00935$; it also clears the PCA
family, whose draws nevertheless reach $q_{95}\cos=.574$.  The sign-randomized
mean comparison remains a non-pass under the frozen conditional randomization
model.  The strict-tail statistic $z_7=-1.500$ fails
all three families (adjusted $p=.969,.872,.794$), so the complete
intersection--union verdict is negative---but per Section~\ref{sec:gates} the
tail component's power is uncalibrated, so this is a failure to establish, not
evidence of absence.  A separate sign-reversal check passes: the exact negation
moves the effect to $-3.898$.

\paragraph{Takeaway.}
CAA's mean effect clears a random-direction baseline but not the structured
conditional comparator, and the protected tail fails all three families.  The
alignment-leakage diagnostic explains why the structured reference is difficult
to separate from the observed vector without reversing the frozen decision or
declaring the randomization model invalid.  Section~\ref{sec:dose} asks what the
geometric baseline and the retained-alignment profile establish together.

\section{Measured effect is associated with alignment at a matched KL budget}
\label{sec:dose}

The 960 reconstructed directions come with their measured effects, giving a
post-hoc alignment--effect profile at a fixed functional budget, measured on
real steered forward passes rather than implanted.  Because cosine was observed,
not experimentally assigned, the profile is descriptive rather than causal.

Alignment is strongly associated with effect: Spearman $\rho(\cos,\Delta b)$ is
$+.938$ within the sign-randomized family and $+.843$ within the PCA family
(within the isotropic family, $+.157$, because its cosine range is only
$\pm.024$---there is nothing to correlate with).  Pooled over all 960 draws,
$\rho=+.785$.

\begin{table}[ht]
\caption{Descriptive alignment--effect profile, all 960 KL-matched directions
pooled into signed-cosine bins, with the observed direction at $\cos=1$.
Family-specific associations are reported in the text.}
\label{tab:doseresp}
\centering
\small
\begin{tabular}{rrrr@{\hspace{1.4em}}rrrr}
\toprule
$\cos$ & $n$ & $\Delta b$ & \% obs. & $\cos$ & $n$ & $\Delta b$ & \% obs. \\
\midrule
$-0.883$ & 31 & $-3.404$ & $-80\%$ &\quad $+0.299$ & 82 & $+3.072$ & $\mathbf{+73\%}$ \\
$-0.491$ & 60 & $-0.704$ & $-17\%$ &\quad $+0.491$ & 63 & $+3.596$ & $+85\%$ \\
$-0.123$ & 102 & $+1.089$ & $+26\%$ &\quad $+0.689$ & 55 & $+3.842$ & $+91\%$ \\
$-0.001$ & 374 & $+1.230$ & $\mathbf{+29\%}$ &\quad $+0.900$ & 26 & $+4.125$ & $+97\%$ \\
$+0.120$ & 53 & $+2.305$ & $+54\%$ &\quad $+1.000$ & 1 & $+4.232$ & $100\%$ \\
\bottomrule
\end{tabular}
\end{table}

The isotropic family alone---a near-orthogonal geometric baseline, mean
$\cos=-.0007$---has
mean effect $1.120$, i.e.\ $26.5\%$ of the observed effect.  Moving from cosine
$0$ to $0.30$ buys another 44 points of the effect; the remaining 70 points of
alignment buy only 27 more.

\paragraph{Consequence for specificity claims.}
A random-direction control asks whether the observed direction beats the
cosine-$\approx0$ region of this profile.  That comparison is cleared, while
directions near cosine $.30$ also produce substantial measured effects and
clear the isotropic $q_{95}$ of $2.349$.  Thus the baseline supports
outperformance of near-orthogonal random directions, but not uniqueness of the
observed construction.  The pooled profile is relatively flat at high positive
cosine, where a strong directional-identification claim would require the most
separation.

\paragraph{What the curve does establish.}
The measured effect is direction-dependent and signed: the exact negation lands
at $-3.898$, the $\cos\approx-0.88$ bin at $-3.404$, and the curve is monotone
across its whole range.  CAA is not doing nothing, and it is not a generic
perturbation effect: at the same KL budget, direction sign flips the outcome.
The supported statement is quantitative rather than binary: within this cell,
substantial effect is observed for directions that are less aligned with CAA,
while the observed vector remains the largest point in the pooled profile.

\paragraph{Caveats.}
Table~\ref{tab:doseresp} pools three families that differ in more than
alignment, so the binned means are descriptive of the pooled population rather
than a controlled alignment experiment; the within-family Spearman coefficients
are family-stratified rather than causal.  The ladder uses $B=100$ per family under a
$q_{95}$ rule while the confirmatory audit uses $B=320$ under Holm, so
Tables~\ref{tab:ladder} and \ref{tab:sc1} are not on the same decision scale.
All of this is one behaviour, one layer, one model.

\section{Transfer, polarity, and what they add}
\label{sec:other}

Two further frozen studies test whether the audited effect survives new data and
a polarity contrast.  Full protocols, prose, and diagnostics are in
Appendices~\ref{app:transfer} and \ref{app:polarity}; the decisions are here.

\begin{table}[ht]
\caption{Independent-bank results in three selected model-layer cells.  Entries
are $D_{95}$ with the $99\%$ lower confidence bound in parentheses.  Cells are
not pooled.}
\label{tab:s3}
\centering
\small
\begin{tabular}{lrrl}
\toprule
Model & Margin & Accuracy & Verdict \\
\midrule
Qwen3-14B        & 0.3945 (0.2606)   & 0.0010 ($-0.0029$) & Margin only \\
DeepSeek-V2-Lite & $-0.0114$ ($-0.0292$) & $-0.0146$ ($-0.0264$) & No replication \\
Gemma-3-27B      & $-0.5966$ ($-0.8479$) & 0.0039 ($-0.0020$) & No replication \\
\bottomrule
\end{tabular}
\end{table}

On a fresh FEVER-derived bank of 1{,}024 confirmation rows, evaluated in three
selected model-layer cells under one frozen protocol
\citep{thorne2018fever,yang2025qwen3,deepseek2024v2,gemmateam2025gemma3}, only
Qwen retains continuous margin transport, forced-choice accuracy transfers in
none of the three cells, and no cell passes the complete specificity gate.
$D_{95}$ compares the sign-randomized and isotropic null tails, so after
Section~\ref{sec:null} it should be read as a measure of how much construction
information the sign-randomized reference retains relative to the isotropic
reference on this bank, not as a standalone measure of efficacy or randomization
validity.

\begin{table}[ht]
\caption{Polarity-mirrored transfer results.  Effects are observed minus base
score with $99\%$ lower confidence bounds in parentheses.  The last three gates
require both polarity cells to pass.}
\label{tab:polarity}
\centering
\scriptsize
\setlength{\tabcolsep}{2.5pt}
\begin{tabular}{llrrcccc}
\toprule
Model & Endpoint & Wrong user & Correct user &
Wrong+NI & Signed & Tail order & Specificity \\
\midrule
Qwen3-14B        & choice       & 1.8553 (1.2160) & 3.0629 (2.7908)   & pass & pass & pass & fail \\
Qwen3-14B        & continuation & .1600 (.1391)   & $-.1251$ ($-.1444$) & fail & fail & pass & fail \\
DeepSeek-V2-Lite & choice       & .3127 (.0923)   & $-.2918$ ($-.5224$) & fail & fail & fail & fail \\
DeepSeek-V2-Lite & continuation & .0868 (.0757)   & $-.1304$ ($-.1519$) & fail & fail & fail & fail \\
\bottomrule
\end{tabular}
\end{table}

Replaying the same frozen directions on an output-unseen 64-fact OpenBookQA bank
with a wrong-user prompt and a correct-user mirror per fact, all four
model-endpoint cells raise the wrong-user score---the metric a paper reporting
this steering vector would naturally report---but only the Qwen choice-score
cell also preserves correct-user performance at the frozen $-.10$ noninferiority
margin.  The other three raise the wrong-user score while lowering the
correct-user score at both endpoints, a pattern consistent with generic
opposition to the user rather than factual correction.  That conclusion rests on
the polarity contrast and the exact-negation check, not on any null family; the
specificity column remains conditional on the frozen exchangeability model, and
the isotropic family is the only purely geometric comparator.

\paragraph{Human-calibrated free generation.}
Three raters independently completed 768 direct-arm and 600 stratified-null
rows (4{,}104 response ratings).  On varying dimensions, Krippendorff $\alpha$
is $.985$--$1.000$; truth stance is unanimous, while 15 user-relation rows split
$2$--$1$ and none needs a fourth rater.  Table~\ref{tab:human-e3} applies the
preregistered human-core O1/O2 tests.

\begin{table}[ht]
\caption{Human-core E3: observed-minus-base correct-fact rate (one-sided $99\%$
lower bound).  O1 requires wrong-user $>0$; O2 requires correct-user $>-.10$.}
\label{tab:human-e3}
\centering
\small
\begin{tabular}{lrrccc}
\toprule
Model & Wrong user & Correct user & O1 & O2 & Joint \\
\midrule
Qwen3-14B        & $.016$ ($-.094$) & $.000$ ($.000$)  & fail & pass & fail \\
DeepSeek-V2-Lite & $.172$ ($.047$)  & $.031$ ($-.063$) & pass & pass & pass \\
\bottomrule
\end{tabular}
\end{table}

DeepSeek passes both gates; Qwen fails O1.  The locked Gemma judge also fails
calibration (class-macro F1 $.562<.80$; minimum supported primary-class F1
$0<.70$), so human-core O1/O2 remain confirmatory but judge-wide O3/O4 are
descriptive.  This endpoint disagreement is an audit finding, not a general
judge benchmark.

\section{Limitations}
\label{sec:limits}

\paragraph{The central measurement is post-hoc.}
It was not preregistered; it was made possible by the frozen seed and the
retained per-draw ledger, and it is exactly reproducible from released code.  It
changes the interpretation of a preregistered verdict, not the verdict itself,
which is retained unaltered in Appendix~\ref{app:ledger}.
Equation~\ref{eq:A} is a first-order concentration argument validated against 21
simulated cells ($r=.979$, mean absolute error $.075$) rather than proved; it
will degrade when $\|d_i\|$ varies strongly across pairs, since it is the
denominator's concentration that makes $\sqrt n$ cancel.  An earlier draft
proposed a signal-to-noise heuristic for the same quantity; it agreed with the
audited cell by coincidence, fails across the panel ($r=.260$), and is
superseded.  Both are computed in the released script so the correction is
auditable.

\paragraph{Scope.}
The alignment-leakage diagnostic spans 21 cells from three stored banks and two
related Qwen model series, but the alignment--effect profile does not:
Table~\ref{tab:doseresp} is one behaviour, one layer, one model, one
construction, and extending it requires steered forward passes.  Whether the
profile transfers is the most important question we leave open.  The
primary evidence throughout concerns one anti-sycophancy/truth-consistency
construction with CAA as the real-direction source; accuracy is a thresholded
view of the same margin, not an independent endpoint; each architecture
contributes one checkpoint-layer cell, so between-cell differences are
setting-conditional rather than causal model-family effects.  There is no Llama
result, and the Gemma cell reuses the locked bank and was registered after
partial results.  The 21-cell analysis is therefore not independent
model-family transfer.

\paragraph{Gate power and conditionality.}
The 512-item configuration retains a strict protected-tail gate with limited
tail power; $B=999$ improves Monte Carlo resolution, not confirmation-sample
power.  The implant ladder calibrates only the mean component, so the complete
gate's sensitivity remains unknown, and the Section~\ref{sec:positive} nominal
passes are sensitive to a conservative $\Gamma=1.10$ envelope.  Matched KL is
conditional on the neutral reference distribution, token horizon, and model
setting.  The PCA null family reaches $q_{95}\cos=.574$ with $\vcaa$, so it is
intermediate in retained alignment between the near-orthogonal geometric
baseline and the sign-randomized family.
Answer ordering and source are constant in the construction bank and cannot be
tested as exchangeability strata.

\paragraph{Open-generation scope and provenance.}
The frozen 64-token cap saturates $99.996\%$ of Qwen and $97.736\%$ of DeepSeek
outputs; 44/1{,}368 rows are invalid and unresolved.  Agreement is partly
low-entropy: all rows are relevant and 1{,}367 are coherent; alpha is undefined
for zero-variance dimensions.  The human result covers two models and 64
mirrored facts, not long-form generation; judge-wide O3/O4 remain descriptive.
Rater-ID correction and hash reconstruction appear in Appendix~\ref{app:integrity}.

\section{Conclusion}

Specificity audits require both a frozen decision rule and an account of what a
comparator preserves.  At matched off-target KL, isotropic directions average
$26.5\%$ of the observed effect, while sign randomization retains substantial
alignment ($\sd(\cos)=.577$; $P(\cos\ge.5)=.253$).  Across 21 stored
model-layer-bank cells, alignment leakage ranges from $.440$ to $.921$.  These
measurements narrow the structured estimand; they do not themselves invalidate
conditional randomization inference.

The frozen Qwen complete gate remains negative: the mean does not clear sign
randomization and the protected tail fails every family.  This is failure to
establish the intersection--union claim, not evidence of no effect.  Conditional
$p$-values still require sign exchangeability, while $A$ and the cosine
distribution expose geometric discriminability.  Matched KL fixes an operating
point rather than a dose; its approximately quadratic scaling does not justify
outcome extrapolation.

The remaining evidence is deliberately mixed: accuracy transfers in no selected
cell, human-rated generation supports factual correction in DeepSeek but not
Qwen, and the automatic judge fails calibration.  Prospective controls show the
complete gate is passable, while $\Gamma=1.10$ marks its nominal passes as
sensitivity-boundary results.  \method{} keeps efficacy, polarity, specificity,
transfer, and semantic claims separately auditable.

\bibliographystyle{plainnat}
\bibliography{references}

\appendix

\subsubsection*{Ethics statement}

This work audits the reliability of activation steering rather than proposing a
deployment mechanism.  Steering is dual use: the same intervention can aid
controlled analysis or induce unwanted behaviour.  We report failed gates and
model-conditional effects, avoid presenting the method as a safety guarantee,
and release diagnostic code rather than model weights.  Transfer banks derive
from FEVER and OpenBookQA.  The reported three-rater activity used anonymized
model responses; no participant attributes or personal data are analysed.
Polarity mirroring changes the asserted user stance but introduces no personal
or sensitive data.

\subsubsection*{Reproducibility statement}

The project reproducibility archive contains preregistrations, frozen gate definitions,
locked data, analysis code, selected final append-only matrices, split-access
checks, integrity reports, the exploratory open-generation analyzer, the
length-risk audit, the exchangeability sensitivity, the versioned claim audit,
and the author-review provenance audit.  The direction reconstruction, cosine
measurement, alignment--effect analysis, and 21-cell alignment-leakage
diagnostic of Sections~\ref{sec:null} and \ref{sec:dose} are included as
standalone CPU-only scripts that regenerate Tables~\ref{tab:sc1},
\ref{tab:transfer}, and \ref{tab:doseresp} and Figure~\ref{fig:dose} from the
released confirmatory report and activation banks in under a minute; none
requires a GPU or model weights.  A standalone verifier checks hashes, unique
key counts, split flags, matched-budget coverage, and report dependencies.
Direction tensors and the largest ledgers are omitted from the size-limited
conference upload; their hashes, completion records, and integrity audits are
included, and the full ledgers accompany the archival release.  Failed,
aborted, and post-lock divergent artifacts are released separately and are not
pooled into their replacements.

\subsubsection*{AI use statement}

Generative AI tools assisted with methodology feedback, code implementation and
debugging, data cleaning and reformatting, semantic prescreening, qualitative
analysis, result summarization, and manuscript drafting and editing.  AI
screening was advisory; the authors remain responsible for the retained data,
facts, frozen gates, numerical claims, integrity reports, and manuscript text.
The authorization-time record states that an author reviewed the retained bank
before model-output access.

\section{The SteerCheck protocol}
\label{app:protocol}

\begin{enumerate}
\item \textbf{Freeze the operating point.}  Neutral prompt bank disjoint from
the behavioural data, token horizon, target off-target KL $\kappa_\star$.
\emph{Catches:} incomparable doses across directions.
\item \textbf{Fit every coefficient blind.}  Bracket-search $c(v)$ against
$\kappa_\star$ on the neutral bank only, for observed and null directions alike.
\emph{Catches:} coefficient tuning that silently optimizes the endpoint.
\item \textbf{Measure the alignment--effect profile.}  Record
$\cos(v,v_{\mathrm{obs}})$ for every evaluated direction, and plot effect
against it.  \emph{Catches:} a specificity claim resting on an empirically flat
region; this is descriptive unless alignment is assigned.
\item \textbf{If you build a construction-matched comparator, audit it.}
Report $A$ and the empirical distribution of
$\cos(v_{\text{null}},v_{\text{obs}})$; no forward pass is required.
\emph{Catches:} retained target alignment that narrows the comparator's
discriminability.
\item \textbf{Decompose the claim}---continuous score change,
observed-versus-null specificity, exact-negation signed transport, discrete
accuracy---per model and per layer, unpooled.  \emph{Catches:} a
metric-conditional effect presented as efficacy.
\item \textbf{Calibrate before concluding negatively.}  Implant
known-alignment directions at the same operating point and report the alignment
at which your statistic detects them.  \emph{Catches:} an underpowered gate
reported as evidence of absence.
\end{enumerate}

Steps 1--3 and 5 are required for a positive claim; 4 and 6 for a negative one.
A standalone verifier checks hashes, unique keys, split flags, matched-budget
coverage, and report dependencies.

\paragraph{Estimator and decision definitions.}
For confirmation item $j$ with score $s_j$,
\begin{equation}
\delta_j(v)=s_j(\theta,c(v)v)-s_j(\theta,0),\quad
T_{\mathrm{mean}}(v)=\frac1m\sum_{j=1}^{m}\delta_j(v),\quad
T_{\mathrm{tail}}(v)=\delta_{(7)}(v),
\end{equation}
\begin{equation}
p_{f,r}=\frac{1+\sum_{b=1}^{B}\mathbf{1}\{T_r(v_f^{(b)})\ge T_r(\vcaa)\}}{B+1},
\qquad
\operatorname{PASS}_{\mathrm{IUT}}=
\bigwedge_f\bigwedge_{r\in\{\mathrm{mean},\mathrm{tail}\}}
[\widetilde p_{f,r}<\alpha],
\end{equation}
with $\widetilde p_{f,r}$ the frozen within-statistic Holm adjustment.  For
independent-bank null ordering we report
$D_{95}=q_{95}^{\mathrm{shuf}}-q_{95}^{\mathrm{iso}}$ with a $99\%$ lower
confidence bound from 20{,}000 frozen bootstrap resamples.

\section{Independent-bank transfer study}
\label{app:transfer}

We constructed a new bank from FEVER \citep{thorne2018fever}: 512 construction
rows, 1{,}024 confirmation rows, and 256 neutral-KL prompts, balanced across
target sign, answer position, source label, and eight seeded prompt templates,
with construction and confirmation claims disjoint and all 1{,}536 rows screened
and reviewed before locking.  Qwen3-14B (layer 28) and DeepSeek-V2-Lite-Chat
(layer 18) were evaluated under one frozen protocol
\citep{yang2025qwen3,deepseek2024v2}; Gemma-3-27B (layer 43) was preregistered
as an extension on the same locked bank after partial results for the other two
were known \citep{gemmateam2025gemma3}.  Each model produced 962 unique
records---observed, negation, and 320 per null family---all KL matched and
passing split, key, and manifest audits.

Qwen passes continuous-margin null ordering and signed margin transport:
observed margin change $0.3655$ ($99\%$ LCB $0.3153$), negated direction
$-0.8986$ ($99\%$ UCB $-0.8056$).  Accuracy transport fails, with exact
one-sided $p$-values $0.6875$ and $0.8125$.  The selected DeepSeek cell passes
neither metric.  The selected Gemma cell fails both: its observed margin moves
adversely ($-0.1340$, $99\%$ LCB $-0.1981$), although the negated direction
moves as expected, so the joint signed gate fails.

\paragraph{A note on $D_{95}$ after Section~\ref{sec:null}.}
This statistic compares the sign-randomized and isotropic reference tails.  It
should be read as a measure of how much construction information the
sign-randomized family retains relative to the geometric baseline on this bank,
not as evidence about steering efficacy or randomization validity.  It was
always described as not an efficacy test; Section~\ref{sec:null} makes the
retained-alignment interpretation explicit.

Accuracy passes in zero of three cells, and no cell passes the complete
specificity gate.  Because Gemma is a sequential same-bank extension, it
broadens the evaluated settings but is not an independent new-data replication.
The earlier single-metric precursor study, which reversed both expected signs,
is retained as a failed preregistered study.

\section{Polarity-mirrored study}
\label{app:polarity}

We replayed the frozen Qwen3-14B and DeepSeek-V2-Lite directions on an
output-unseen 64-fact OpenBookQA bank, each fact producing a wrong-user prompt
and a correct-user mirror.  The 128 prompts were evaluated under base, observed,
exact-negation, and 320 directions per null family, yielding 123{,}264 records
per model and endpoint; all ledgers pass the append-only key, response-hash,
authorization, and coverage audit.  The \emph{choice-score} endpoint scores
complete response choices; the \emph{continuation} endpoint scores
length-normalized correct versus incorrect continuations.

All four cells pass the wrong-user correction-score gate---the metric a paper
reporting this steering vector would most naturally report.  Only the Qwen
choice-score cell also passes correct-user noninferiority at the frozen margin
of $-.10$, and even there the pass is score-level: wrong-user accuracy moves
from $.7969$ at base to $.7812$ under the observed direction, while correct-user
accuracy remains $.9844$.  At the continuation endpoint, Qwen increases the
correct-continuation score for wrong users but violates correct-user
noninferiority and reduces the positive-margin fraction from $.3438$ to $.1094$.
DeepSeek shows the same asymmetry at both endpoints.  \textbf{These three
asymmetric cells are consistent with generic opposition to the user rather than
factual correction}---a conclusion that rests on the polarity contrast and the
exact-negation check, not on any null family.

The specificity column remains conditional on within-pair exchangeability: the Qwen
choice-score cell's adjusted $p$-values are $.00935$ against isotropic, $.04984$
against PCA-subspace, and $.08100$ against sign randomization.  The isotropic
family is the only purely geometric comparator; the other two deliberately
retain progressively more construction geometry.

\section{Evidence binding and verification}
\label{app:evidence}

Every confirmatory result is bound to an append-only ledger with a content hash
written before the corresponding output was accessed.  The top-level frozen
matrix report for the positive-control study is
\texttt{4ae18aab\ldots beaa1736}; each of its four cell-specific ledgers
contains 2{,}999 unique successful keys, unpooled.  Construction geometry
selected Qwen layers 20/28 and DeepSeek layers 24/18.  A prior synthetic stage
audited four 962-key full synthetic cells and four 98-key probes before
selecting $B=999$, which improves finite Monte Carlo resolution without
increasing the 512 confirmation items that power the protected 26th-order
statistic.

The complete hash manifest---bank lock, geometry lock, code lock, authorization,
execution lock, and per-cell raw/status/metadata/report digests---is released in
the project archive as \texttt{MANIFEST.json} with a standalone verifier.  Retained
allocation failures, container scheduling failures, a fixed-horizon traceback, a
cancelled pre-result chain, and strict-resume histories are preserved separately
from the authoritative ledgers.

\paragraph{Reconstruction provenance.}
The 960 confirmatory null directions are regenerated with NumPy's
\texttt{default\_rng}, using seed \texttt{SEED + 1 + seed\_offset}, with
\texttt{SEED = 20260717} and \texttt{seed\_offset = 0}.  Draws are consumed in the frozen
family order (isotropic, PCA, sign-randomized), 320 draws each, against the
locked vectors file.  The reconstruction is validated by recomputing each draw's
$\|v\|$ from the ledger's \texttt{perturbation\_norm / c}; agreement is within
$1.7\times10^{-16}$ relative error for all 320 sign-randomized draws, which is
exact in double precision.

\section{Complete evidence ledger}
\label{app:ledger}

Aborted and superseded runs are retained as separate evidence and are not pooled
into their replacements.  The confirmatory verdict of the discovery audit is
retained here as originally frozen; Section~\ref{sec:null} refines what the
structured comparator can establish without changing its numbers.

\begin{table}[ht]
\caption{Authoritative experiment ledger.}
\label{tab:ledgertab}
\centering
\scriptsize
\setlength{\tabcolsep}{3pt}
\begin{tabular}{lrp{7.2cm}}
\toprule
Stage & Scale & Frozen outcome \\
\midrule
Discovery screen & --- & Qwen3 anti-sycophancy passes; benign compliance fails monotonicity. \\
Monte Carlo power gate & 199/299 & $B=199$ aborts for power $.887<.90$; $B=299$ passes with power $.947$. \\
First GPU implementation & smoke & First implementation aborts; repaired implementation recovers 6/6 synthetic signatures. \\
Reachability sweep & 2{,}196 & No authoritative common budget. \\
Metric-panel sweep & 6{,}588 & Metric- and panel-dependent labels; operational labels are not mechanism explanations. \\
Cross-model reachability & 59{,}292 & Nine cells reachable and side-channel dominant, but anchors do not reproduce; label-level causal-path claim withdrawn. \\
LoRA learned directions & 6{,}588 & Six of six reachable; zero of six recover a mechanism signature. \\
Dose extension & 6{,}588 & Score shifts grow but answer flips remain rare; the only flip-floor hit costs KL $64.66$. \\
KL--coefficient scaling & 12 fits & Quadratic law holds; efficacy per $\sqrt{\kappa}$ not constant across radius. \\
Graft detection rule & 24 cases & Rule v1 sensitivity 6/18 fails; v2 detects 18/18 non-null and rejects 6/6 null cases on the same grafts. \\
Single-metric precursor & 962 & Nonreplication with both expected signs reversed; retained as a failed preregistered study. \\
Discovery audit (confirmatory) & 962 & $\Delta b=4.2321$; iso and PCA mean pass, sign-randomized fails, $z_7$ fails all three. \textbf{Structured-comparator interpretation refined in Section~\ref{sec:null}; verdict unchanged.} \\
Independent-bank study & $3\times962$ & Qwen margin-only; DeepSeek and Gemma negative; accuracy passes 0/3. \\
Exchangeability replay & 640 nulls & Exact seed/hash replay; balanced subsets retain all four $\alpha=.01$ decisions. \\
Polarity-mirrored study & $4\times123{,}264$ & Four wrong-user score passes, one joint pass, zero four-way specificity passes. \\
Open-generation labels & 246{,}528 & Automatic labels complete and integrity-audited; reported as exploratory only. \\
Positive-control study & $4\times2{,}999$ & Both language controls pass the complete IUT; Qwen detox does not, DeepSeek detox does. \\
\textbf{Alignment-leakage measurement} & 960 dirs & \textbf{Post-hoc CPU reconstruction; sign-randomized $\sd(\cos)=.577$, $P(\cos\ge.9)=.047$; frozen randomization verdict unchanged.} \\
\textbf{Alignment-leakage panel} & 21 cells & \textbf{CPU simulation over three construction banks $\times$ seven layers in two related Qwen series; $\sd(\cos)\in[.440,.921]$; $A$ predicts at $r=.979$.} \\
\bottomrule
\end{tabular}
\end{table}

\FloatBarrier
\section{Supporting diagnostics}
\label{app:diag}

\begin{figure}[ht]
\centering
\includegraphics[width=\linewidth,height=2.5in,keepaspectratio]{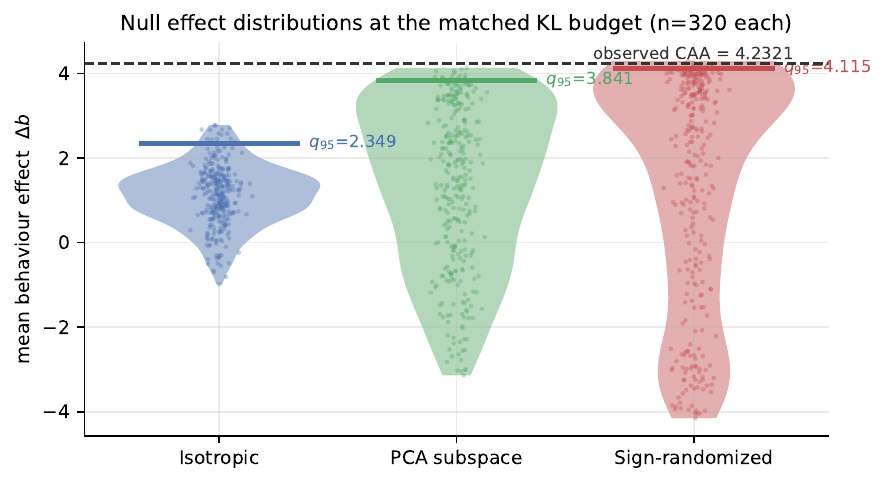}
\caption{Null effect distributions at the matched KL budget, 320 draws per
family, with $q_{95}$ markers and the observed CAA line.  This is the comparison
Table~\ref{tab:sc1} tests.  Read together with Table~\ref{tab:sc1}, the
rightward shift of the sign-randomized family reflects its retained alignment
with $\vcaa$, not a property of label-free directions.}
\label{fig:nulls}
\end{figure}

\begin{table}[ht]
\caption{Relationship to reliability-oriented steering evaluations.  ``Open''
denotes open-ended generation; ``matched null'' denotes a null produced by the
same contrast construction rather than an arbitrary random vector.}
\label{tab:related}
\centering
\scriptsize
\setlength{\tabcolsep}{2.5pt}
\begin{tabular}{lccccc}
\toprule
Work & Context & Open & Geom. & Variation & Audit/calibration \\
\midrule
\citet{pres2024reliable}      & yes & yes & -- & -- & -- \\
\citet{braun2025unreliability}& yes & --  & yes & -- & -- \\
\citet{dasilva2025steering}   & yes & --  & -- & models & -- \\
\citet{ye2026sources}         & yes & --  & -- & sources & -- \\
\citet{herbster2026openended} & yes & yes & coherence & -- & -- \\
\citet{goyal2026specificity}  & gen./ctrl. & yes & -- & -- & specificity tests \\
\textbf{SteerCheck (ours)}     & forced + mirrored & explor. & KL budget & fixed CAA &
\textbf{alignment profile + leakage audit} \\
\bottomrule
\end{tabular}
\end{table}

\paragraph{The sign-exchangeability assumption.}
The sign-randomized family's null interpretation assumes that, absent the
intended directional association, the orientation within every pair is
exchangeable.  Construction and confirmation banks are balanced by truth label
and prompt template.  In the construction bank, response ordering is fixed by
design (the positive target begins ``No'' and the negative target ``Yes''), and
the source dataset and split are constant, so those variables cannot serve as
empirical exchangeability strata.  It is therefore not an unconditional exact
test of semantic label independence.  Exchangeability remains the validity
condition.  Section~\ref{sec:null} measures a separate property---internal
agreement and the resulting alignment leakage---that governs geometric
discriminability, not validity.

\paragraph{Implant-ladder cross-check.}
An independent known-alignment implant ladder, built against the measured
behaviour \emph{gradient} rather than against $\vcaa$, gives a consistent
picture at the same operating point.

\begin{table}[ht]
\caption{Mean-component response to implanted alignment $t$ with the behaviour
gradient, at the frozen KL budget.}
\label{tab:ladder}
\centering
\small
\begin{tabular}{lrrrrrrr}
\toprule
Alignment $t$ & .20 & .15 & .10 & .07 & .05 & .03 & .02 \\
\midrule
Mean $\Delta b$ & 4.271 & 3.938 & 3.423 & 2.923 & 2.499 & 2.004 & 1.702 \\
Detected ($q_{95}$ rule) & yes & no & no & no & no & no & no \\
\bottomrule
\end{tabular}
\end{table}

A ten-fold increase in gradient alignment, from $t=.02$ to $t=.20$, buys a
2.5-fold increase in effect---the same saturation, on a different axis.  The
real CAA direction has measured gradient alignment $.1539$, for which the ladder
predicts $3.94$--$3.96$ against an observed $4.232$.  The two axes are not
interchangeable---cosine with $\vcaa$ is not cosine with $\nabla s$, and CAA
itself is only $.15$ aligned with the gradient---so we report them separately
without treating either profile as a causal intervention on alignment.

\paragraph{Re-scoring the ladder under the gate as frozen.}
Table~\ref{tab:ladder} scores detection under a $q_{95}$ rule, whereas the
confirmatory verdict uses Holm $\alpha=.01$, which at $B=320$ requires zero
exceedances.  Against the isotropic family that threshold is the isotropic
maximum $2.776$, reached between $t=.05$ and $t=.07$; a direction of CAA's
gradient alignment clears it.  We note this because an earlier draft of this
paper reported a detection floor derived against the sign-randomized family,
which Section~\ref{sec:null} shows is a leakage-affected conditional comparator
rather than an alignment-free reference.

\paragraph{Exchangeability sensitivity.}
A CPU-only sensitivity was frozen before inspection.  All 640 structured
directions reconstruct exactly from their recorded seeds and hashes---the same
property Section~\ref{sec:null} exploits.  Restricting to directions whose 16
truth-by-template cells each have absolute mean sign at most $.375$ retains
225/320 Qwen and 236/320 DeepSeek nulls; raw $p$-values change from $.1807$ to
$.1549$ (Qwen margin), $.4673$ to $.4690$ (Qwen accuracy), $.7882$ to $.8059$
(DeepSeek margin), $.6885$ to $.7089$ (DeepSeek accuracy), and all four
$\alpha=.01$ decisions remain negative.  Nineteen of 30 predeclared
activation-geometry tests flag truth/template-linked heterogeneity after Holm
correction.  This supports decision stability under the specified balance
envelope, not exchangeability itself.  The alignment-leakage audit is a separate
diagnostic and cannot establish or refute exchangeability.

\subsubsection*{Human-calibrated open-generation evidence}
The blinded packet contains all 768 base/observed/negated responses and 600
frozen-stratified null responses.  Majority vote is defined per dimension.
Across 1{,}368 rows, truth stance, coherence, invalidity, refusal, relevance,
and repetition are unanimous; 15 user-relation rows split $2$--$1$, and no row
has a three-way disagreement.  Krippendorff $\alpha$ is $1.000$ for truth
stance, coherence, and invalidity and $.985$ for user relation; it is undefined
for the three zero-variance dimensions.  Consensus truth counts are 1{,}141
correct, 146 false, 34 mixed, and 47 unresolved.  The frozen paired bootstrap
(20{,}000 resamples, seed 20260803) yields Table~\ref{tab:human-e3}.

\paragraph{Automatic-judge calibration and exploratory nulls.}
Against human consensus, the locked \texttt{gemma-3-27b-it} parser-v1.1 judge
has overall class-macro F1 $.562$, truth-stance macro-F1 $.405$, and a minimum
supported primary-class F1 of zero.  It therefore fails both frozen thresholds.
Its 246{,}528 complete labels remain exploratory: observed joint semantic
quality is $.773$ against a null-family median of $.812$ for Qwen, and $.859$
against $.906$, $.906$, and $.891$ for DeepSeek.  The 64-token cap saturates
$99.996\%$ of Qwen and $97.736\%$ of DeepSeek outputs, and arm lengths differ by
up to $1.32$ tokens.

\paragraph{Disclosed correction.}
The sealed first-version analyzer correctly separated the wrong-user and
correct-user gates but pooled polarity for signed transport, null-tail ordering,
and specificity, and omitted preregistered choice-score accuracy.  We retain its
reports and disclose a post-output protocol-conformance correction that
separates those cells without changing any threshold, seed, model, direction,
coefficient, null, or prompt.

\paragraph{Implant construction.}
The frozen known-alignment implant of Table~\ref{tab:ladder} is
\begin{equation}
\widehat g_b=\frac{g_b}{\|g_b\|_2},\qquad
\widehat n_\perp=\frac{n_\perp}{\|n_\perp\|_2},\qquad
\widetilde v(t)=\|\vcaa\|_2\left(t\widehat g_b+\sqrt{1-t^2}\,\widehat n_\perp\right),
\end{equation}
with $\widehat n_\perp^\top\widehat g_b=0$, so $t$ is the exact cosine with the
measured behaviour gradient while the real direction's norm is preserved before
KL matching.  Four seeds per level; a level counts as detected when at least
$75\%$ of seeds clear the $q_{95}$ of all three 100-draw null families.

\paragraph{Layer selection.}
Construction-only leave-one-out coherence selected layer 20 ($0.708$) and 28
($0.403$) for Qwen and 24 ($0.717$) and 18 ($0.402$) for DeepSeek under the
lower-layer tie rule.  The reserve bank was not accessed.

\paragraph{Reachability and metric conditionality.}
The reachability sweep (2{,}196 points) finds no common budget.  At
$\kappa=10^{-5}$, reachability is zero except for PCA ($0.20$--$0.30$); at radius
$0.10$, neutral KL $7.8\times10^{-8}$ coexists with a fit-head effect near
$70.72$.  The metric-panel sweep (6{,}588 points) finds two metric-dependent and
four panel-dependent cases, but no locally invisible case.

\paragraph{Cross-model reachability and learned directions.}
59{,}292 records over three layers each of Qwen2.5, Qwen3, and DeepSeek: all
nine cells are broad-context reachable, side-channel dominant, and depth-stable,
but the preregistered Qwen2.5 anchors fail on the independent bank, so the
label-level causal-path claim is withdrawn.  All six Qwen2.5 layer-20 LoRA
directions are reachable, yet none matches its preregistered signature; the
task-score shift grows from roughly 4 to 12 over $\rho\in[.15,.80]$, but only
one instance reaches the $0.10$ flip floor, at $\rho=0.80$ and KL $64.66$.

\section{Integrity, corrections, and retained failures}
\label{app:integrity}

\begin{table}[ht]
\caption{Exploratory automatic-label rates for base, observed, and negated arms.
Joint quality requires a correct-fact stance, appropriate relation to the user's
position, coherence, relevance, no refusal, no invalidity, and no repetition.}
\label{tab:labels}
\centering
\small
\begin{tabular}{llccc}
\toprule
Model & Arms & Truth correct & User relation & Joint quality \\
\midrule
Qwen3-14B        & base/obs./neg. & .898/.867/.914 & .820/.812/.812 & .805/.773/.797 \\
DeepSeek-V2-Lite & base/obs./neg. & .945/.883/.945 & .922/.867/.938 & .922/.859/.930 \\
\bottomrule
\end{tabular}
\end{table}

Output-length saturation and between-arm length imbalance are stated in
Section~\ref{sec:limits}.  The three independently completed rating files contain
1{,}368 unique blind IDs each, no missing or illegal labels, and exact packet-ID
coverage.  All copies initially inherited the demonstration value \texttt{R1}
in the rater-ID column.  After labels were frozen, that field alone was corrected
to \texttt{R1}/\texttt{R2}/\texttt{R3}: reversing the R2 and R3 substitutions
reproduces the pre-correction SHA-256 hashes exactly.  No label changed.  The
source hashes, consensus, calibration report, and byte-level correction check
are included with the manuscript source.  A separate single-author qualitative
face-validity review produced no per-row labels and is not used as calibration
or inter-rater evidence.

\paragraph{Structured-null sensitivity.}
The frozen CPU-only analysis reconstructs all 640 structured-direction hashes.
Maximum joint-cell imbalance is not associated with either outcome after Holm
correction, but 19/30 activation norm/projection/PCA tests satisfy the
predeclared heterogeneity concern rule.  Source and answer-order strata are
degenerate.  These results preserve the original verdict and probe the
exchangeability model; the alignment-leakage analysis in
Section~\ref{sec:null} addresses a separate discriminability axis.

\paragraph{Ledger integrity and retained failures.}
Each final matrix contains 962 rows and 962 unique keys: observed 1, negated 1,
and 320 per null family.  Every row is budget matched.  Confirmation access
occurs only after the signed lock; sealed legacy splits are untouched.  The
archive preserves the $B=199$ power abort, the first GPU implementation failure,
the first ineffective LoRA training set, and the single-metric nonreplication.
Two preregistration defects are retained: the LoRA study's underspecified format
ratio, stable over a $0.1$--$2.0$ sensitivity sweep; and the dose-extension
study's ill-defined one-token coherence guard, whose maximally adverse
counterfactual leaves its gate unchanged.

\paragraph{Three provenance anomalies.}
(i) One report schema emits no self-recorded script hash; the on-disk analysis
script was hashed at read time and recorded in the manifest.  (ii) One
access-time seal was invalidated by a read and is documented rather than
reconstructed.  (iii) The working-tree author-review CSV (SHA-256 prefix
\texttt{dc6406}) differs from the CSV named by the signoff; running the frozen
finalizer over the unchanged review template reconstructs the signed bytes
exactly (prefix \texttt{1aacf1}).  The full-run authorization, execution lock,
lock manifest, and target-bank hashes remain mutually consistent.  None of the
three changes any bound numerical matrix, content hash, split flag, or the
formal authorization chain.

\end{document}